\documentclass[letterpaper, 10 pt, conference]{ieeeconf}  %

\IEEEoverridecommandlockouts                              %

\usepackage[colorinlistoftodos, german, disable]{todonotes} %
\usepackage{url}      %
\usepackage{graphicx} %
\usepackage{epstopdf}
\usepackage{tikz-cd}
\DeclareGraphicsExtensions{.pdf}
\usepackage[utf8]{inputenc}
\usepackage[font=footnotesize]{caption}
\usepackage[font=footnotesize]{subcaption}
\usepackage{hyperref}
\usepackage{siunitx}
\usepackage{amsmath}
\usepackage{ifthen}
\usepackage{amssymb}
\usepackage{tabularx}
\usepackage[USenglish]{babel}
\usepackage[backend=bibtex,bibstyle=ieee,citestyle=numeric-comp,maxbibnames=3,maxcitenames=3,doi=false,url=false,eprint=false]{biblatex}
\usepackage{tikz}

\bibliography{library}

\title{\LARGE \bf
Holistic Graph-based Motion Prediction
}

\author{Daniel Grimm$^{1}$, Philip Schörner$^{1}$, Moritz Dreßler$^{2}$ and J.-Marius Zöllner$^{1,2}$%
\thanks{$^{1}$ FZI Research Center for Information Technology, 76131 Karlsruhe, Germany.
	{\tt\small daniel.grimm, schoerner, zoellner@fzi.de}}%
\thanks{$^{2}$ Karlsruhe Institute of Technology (KIT), Germany.}%
}

\newcommand\copyrighttext{%
  \footnotesize © 2023 IEEE. Personal use of this material is permitted. Permission from IEEE must be obtained for all other uses, in any current or future media, including reprinting/republishing this material for advertising or promotional purposes, creating new collective works, for resale or redistribution to servers or lists, or reuse of any copyrighted component of this work in other works.}
\newcommand\copyrightnotice{%
\begin{tikzpicture}[remember picture,overlay]
\node[anchor=south,yshift=10pt] at (current page.south) {\fbox{\parbox{\dimexpr\textwidth-\fboxsep-\fboxrule\relax}{\copyrighttext}}};
\end{tikzpicture}%
}

\begin{document}

\maketitle

\thispagestyle{empty}
\pagestyle{empty}
\copyrightnotice

\begin{abstract}
Motion prediction for automated vehicles in complex environments is a difficult task that is to be mastered when automated vehicles are to be used in arbitrary situations.
Many factors influence the future motion of traffic participants starting with traffic rules and reaching from the interaction between each other to personal habits of human drivers.
Therefore, we present a novel approach for a graph-based prediction based on a heterogeneous holistic graph representation that combines temporal information, properties and relations between traffic participants as well as relations with static elements such as the road network.
The information is encoded through different types of nodes and edges that both are enriched with arbitrary features.
We evaluated the approach on the INTERACTION and the Argoverse dataset and conducted an informative ablation study to demonstrate the benefit of different types of information for the motion prediction quality.

\end{abstract}

\section{Introduction}
Machine learning has improved in recent years and excels in domains where it is hard to find an explicit mathematical description of the solution.
In autonomous driving machine learning led to great improvements in perception tasks.
However, driving in crowded scenes remains challenging for autonomous vehicles (AVs), mainly because the motion prediction becomes harder due to the increasing number of possible interactions among the traffic participants while paying attention to the road.
This problem is not restricted to autonomous driving and can easily be transferred to other use cases where autonomous systems interact and share their space with humans, e.g., a logistic robot in a warehouse.
In this work we focus on motion prediction for AVs.

In recent works Jia et al. \cite{hdgt} and  Mo et al. \cite{heat} solve the spatio-temporal characteristic of the problem in a two staged fusion approach.
Firstly, dynamic information, i.e. the past trajectory of the traffic participants, is fused over time.
Secondly, information is shared between the traffic participants and the road.
Yuan et al. \cite{agentformer} propose a simultaneous temporal and spacial fusion of the past trajectories using a masked transformer.
This allows to capture the dynamic context at a higher resolution.
However, map information is modelled as a birds eye view image and processed via a convolutional neural network (CNN).
This is not optimal, because each agent should process only surrounding road elements and not the complete map.
Works such as Jia et al. \cite{hdgt} and Liang et al. \cite{lanegcn} model the map as a graph allowing traffic participants to only attend to areas of the map where they are currently driving.
Those approaches use the aforementioned staged fusion approach, which motivates the problem of finding a graph representation that incorporates the spatio-temporal information from traffic participants and the environment.
Thus, we propose a heterogeneous graph for simultaneous attention to past history, other agents' time-discrete trajectory and map information without using pre-fused data.
The contribution of this paper includes:
\begin{itemize}
    \item Holistic heterogeneous graph: Formulating the problem as a graph without pre-fused data makes it possible to capture interactions at a higher resolution.
    \item Modularity: More opportunities to encode expert-knowledge in the graph via edges and their features. The modular construction of the graph also allows for further extensions in the future.
    \item Benchmarking: INTERACTION and Argoverse dataset
\end{itemize}
\begin{figure}[t]
	\centering
	\begin{center}
		\includegraphics[width=\columnwidth]{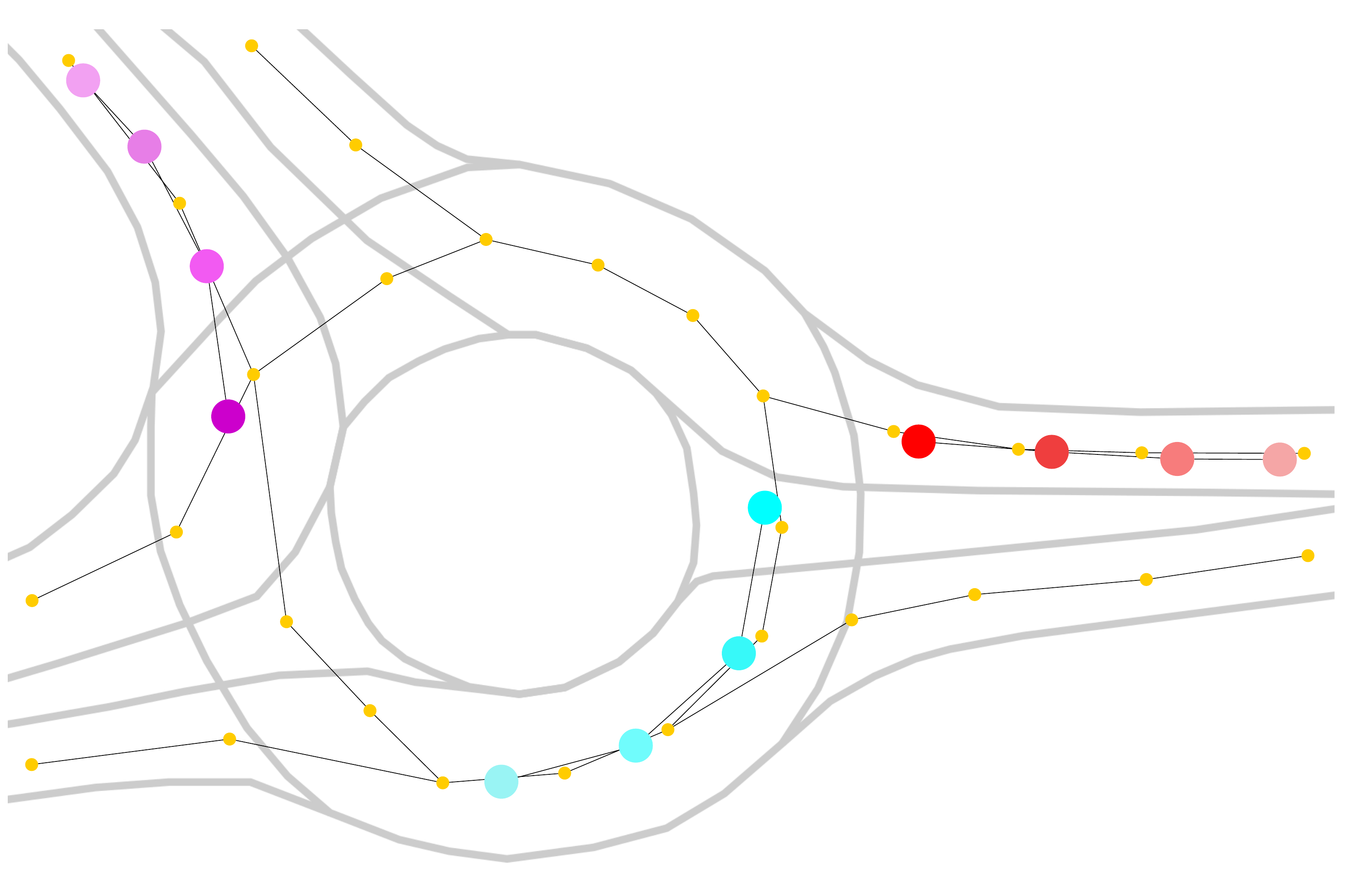}
		\caption{Nodes in the heterogeneous graph. Map-nodes are depicted in yellow. Different colored nodes represent agent-nodes where nodes of the same color belong to the same trajectory. Time context is visualized with color fading. The high definition map (HD-Map) is depicted in light gray for better understanding of the traffic scene.}
		\label{fig:titleimage}
	\end{center}
\end{figure}
\section{Related Work}
Motion prediction is an ongoing research topic in the field of autonomous driving.
In this section we provide an overview regarding graph neural networks (GNN) and motion prediction.
As we are pursuing a learned prediction approach, we are focussing on learning-based approaches for motion prediction.
\begin{figure*}[t]
    \vspace{4pt}
    \centering
    \includegraphics[width=\textwidth]{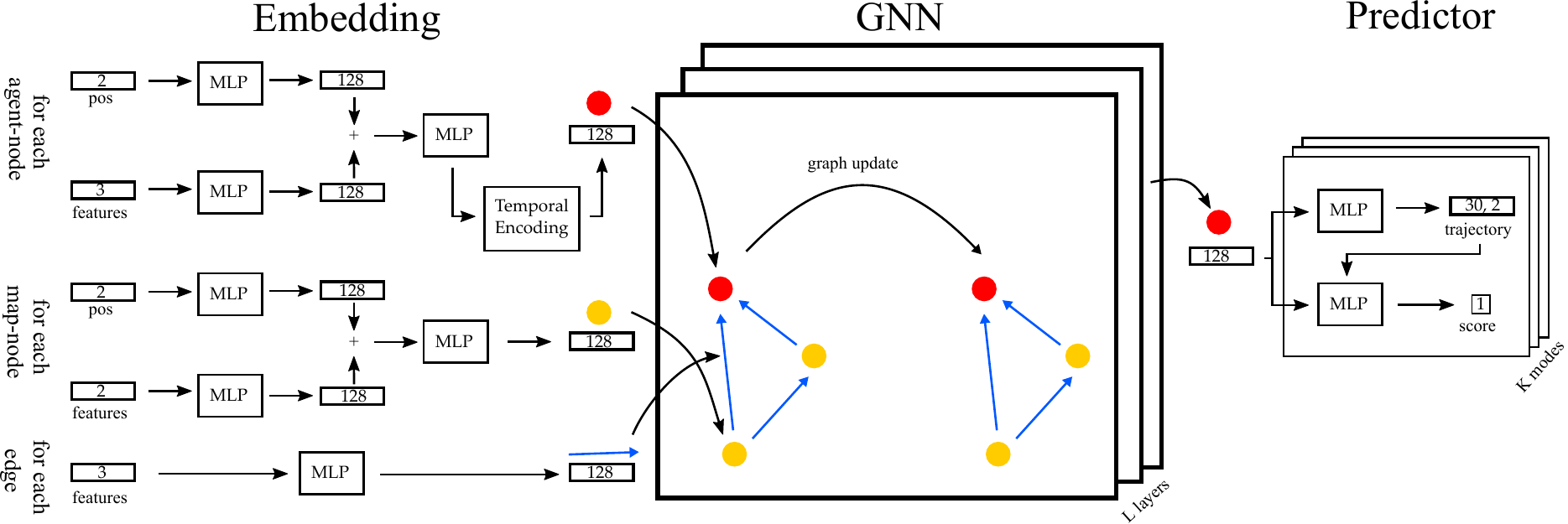}
    \caption{Proposed concept. Inputs are embedded in separate embedding modules. Heterogeneous GNN is used to generate latent representation of all agents in the scene. Prediction head outputs a future trajectory for each agent.}
    \label{fig:concept}
\end{figure*}
\subsection{Graph Neural Networks}
Graph neural networks are used to extract information from data which can be structured in graphs.
For homogeneous graphs, there exist a wide variety of operations to exchange information between nodes, e.g. GCN \cite{gcn}, GraphSAGE \cite{graphsage}, GAT \cite{gat} and Gatv2 \cite{gatv2}, each following the message passing scheme \cite{torch_geometric} to update the nodes in the graph.
Most previous works, such as \cite{gcn}, \cite{gat} focus on homogeneous graphs. 
This is not sufficient in the field of motion prediction, where different entities, e.g., traffic participants and map elements, interact.
Heterogeneous graphs consist of different node and edge types \cite{hetgrahsurvey}.
Wang et al. \cite{han} propose to model attention in a heterogeneous graph in a two stage approach called Node-Level Attention and Semantic-Level Attention.
Hu et al. \cite{hgt} introduces ideas of a Transformer \cite{transformer} in a heterogeneous graph.
The attention matrix is calculated dependent on the edge type and node type. However, edge features are not considered.
\subsection{Motion Prediction}
The task of motion prediction is mostly formulated as a seq2seq problem.
Therefore, early motion prediction models, such as Social LSTM \cite{sociallstm} from Alahi et al., PRECOG \cite{precog} and R2P2 \cite{r2p2} from Rhinehart et al., rely on Recurrent Neural Network (RNN) structures, such as LSTM \cite{lstm} or GRU \cite{gru}.
With the success of CNNs in the domain of image classification, such as Krizhevsky et al. \cite{alexnet} and Simonyan et al. \cite{vgg}, it became possible to use a 2D birds eye view (BEV) image of the street layout in motion prediction.
Hong et al. \cite{rules_of_the_road}, Phan-Minh et al. \cite{covernet} and Djuric et al. \cite{uberprediction} encode a rich representation of the environment including road elements, dynamic context and other traffic participants in the image.
Due to the success of Transformer \cite{transformer} in Natural Language Processing, which is also a seq2seq problem, works, such as Ngiam et al. \cite{scene_transformer} and Mercat et al. \cite{multiheadmultijoint} adopted the attention mechanism for motion prediction.
Yuan et al. \cite{agentformer} combine attention over time and over other agents in one Transformer called AgentFormer.
Attention is done in a fully connected fashion not regarding spatial distance between agents.
To the authors knowledge, VectorNet from Gao et al. \cite{vectornet} and LaneGCN from Liang et al. \cite{lanegcn} were the first models, to use a GNN for motion prediction.
VectorNet uses a local graph to obtain polyline-level features for agent trajectories and lanes.
Afterwards these features are used in a global interaction graph, which is fully connected, undirected and homogeneous.
In contrast to Vectornet, LaneGCN uses a segment of a polyline as a node in their lane graph, hence, capturing the map at a higher resolution.
DenseTNT by Gu et al. \cite{densetnt} adopt Vectornet and split the prediction task into goal prediction and trajectory completion.
The map-nodes in the heterogeneous graph proposed in our model use a similar map representation as LaneGCN.
HEAT from Mo et al. \cite{heat} and HDGT from Jia et al. \cite{hdgt} propose a heterogeneous interaction graph, where the nodes represent higher-level features, such as agent trajectories or lanes.
HEAT constructs the street layout with a CNN from BEV images.
HDGT uses a simplified PointNet \cite{pointnet} to encode lane features from a vectorized format.
Sheng et al. \cite{spatial_temporal_conv}, and Cao et al. \cite{spectgnn} introduce a graph-based spatial-temporal convolution.
Our proposed heterogeneous graph differs from above mentioned works by the differently modelled temporal information.
Instead of fusing temporal information outside of the graph such as HEAT \cite{heat} and HDGT \cite{hdgt}, or using a separate graph for each time-step such as Sheng et al. \cite{spatial_temporal_conv} and Cao et al. \cite{spectgnn}, we combine time variant information, e.g. agent trajectories, in one graph.
Time information is preserved by the usage of a temporal encoding, see Sec.~\ref{sec:graph_embedding}.
To the knowledge of the authors, we are the first to model the whole encoding step in a single graph for the task of motion prediction.
\section{Concept}
The general pipeline is depicted in Fig.~\ref{fig:concept}. The model consists of an embedding part followed by an encoder-decoder structure.
As encoder, we propose a spatio-temporal static heterogeneous graph, which includes encoding the past trajectory as well as social context attention and the encoding of the street layout.
The graph yields a latent feature vector per agent.
The decoder is a normal Multilayer Perceptron (MLP) that outputs multi-modal predictions for each agent in the scene.
We use a scene-centric data representation.
\subsection{Embedding}
\label{sec:graph_embedding}
Firstly, agent-nodes, map-nodes and edge features are embedded to a higher dimension 
$f$ using a set of MLPs each with a linear layer followed by ReLU Activation and Layer-Normalization. A detailed view of the embedding process is depicted in Fig.~\ref{fig:concept}.
In order to represent the timestamp of an agent-node $\pmb{a}_i^t$, a temporal encoding $\pmb{\tau}$ which is similar to the positional encoding in Transformers \cite{transformer} is added to the agent-nodes in the last step of the embedding.
\begin{equation}
    \tau(t, 2i) = \sin{\left(t/10000^{\frac{2i}{f}}\right)}
\end{equation}
\begin{equation}
    \tau(t, 2i+1) = \cos{\left(t/10000^{\frac{2i+1}{f}}\right)}
\end{equation}
\begin{equation}
    \pmb{a}_i^t =  \pmb{W}_{1}\left(\pmb{a}_i^t \mathbin\Vert \pmb{\tau}(t) \right)
    \label{eq:temporal}
\end{equation}
where $\tau(t, 2i)$ and $\tau(t, 2i+1)$ refer to the even resp. odd index of feature dimension in $\pmb{\tau}(t)$ and $t$ refers to id of the time-step, e.g., $t \in\{0,1,...,9\}$ in the INTERACTION dataset.
\begin{figure}[h]
	\centering
    \begin{subfigure}{0.95\columnwidth}
    	\vspace{5pt}
    	\centering
    	\includegraphics[width=0.85\columnwidth]{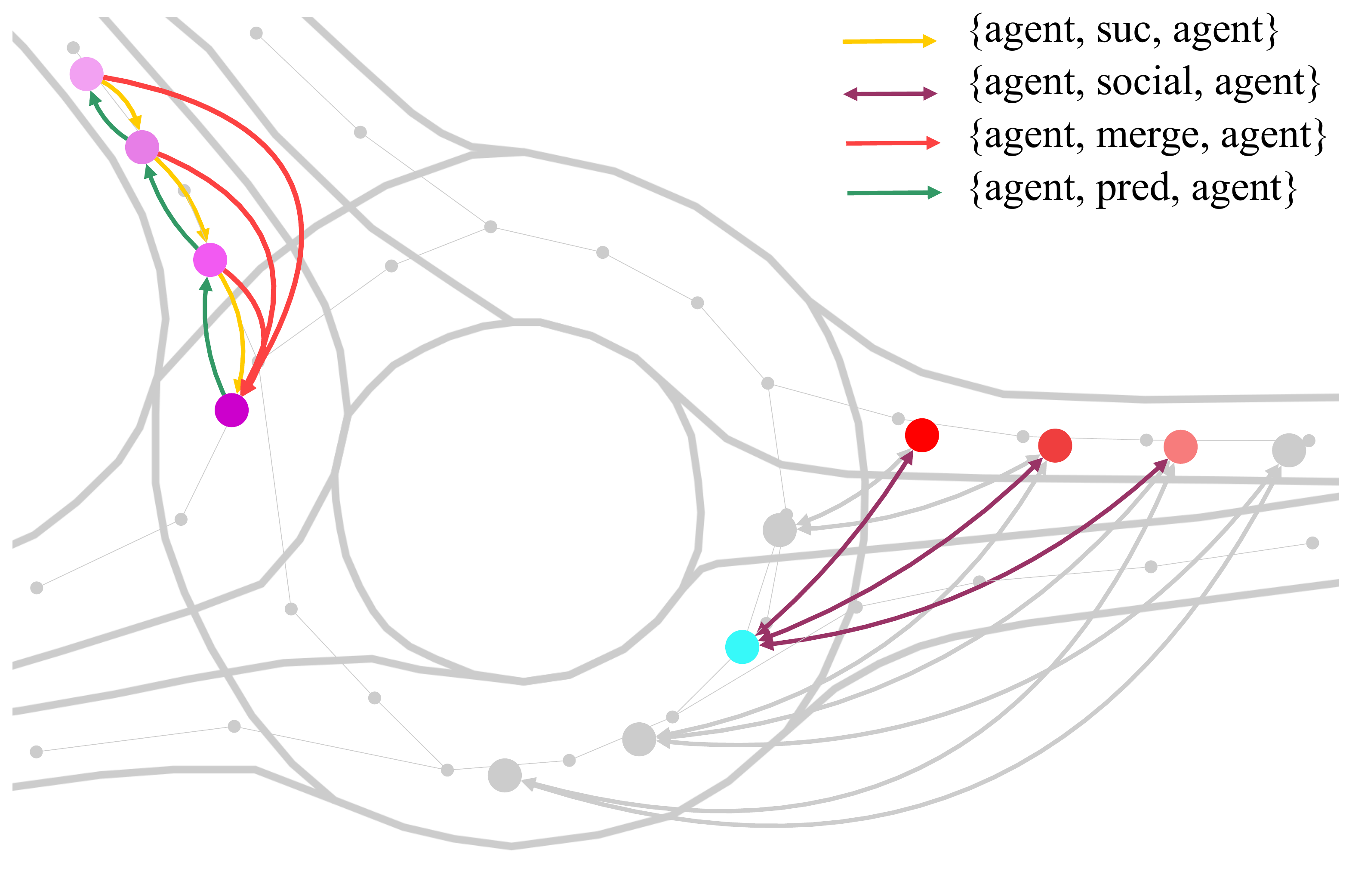}
        \caption{Edges among agent-nodes. Time-step information is presented with color shading. Agent-nodes belonging to one agent trajectory are presented on the left. Social context is visualized on the right.}
  	    \label{fig:agent_edges}
	\end{subfigure}
	\begin{subfigure}{0.95\columnwidth}
    	\vspace{5pt}
    	\centering
    	\includegraphics[width=0.85\columnwidth]{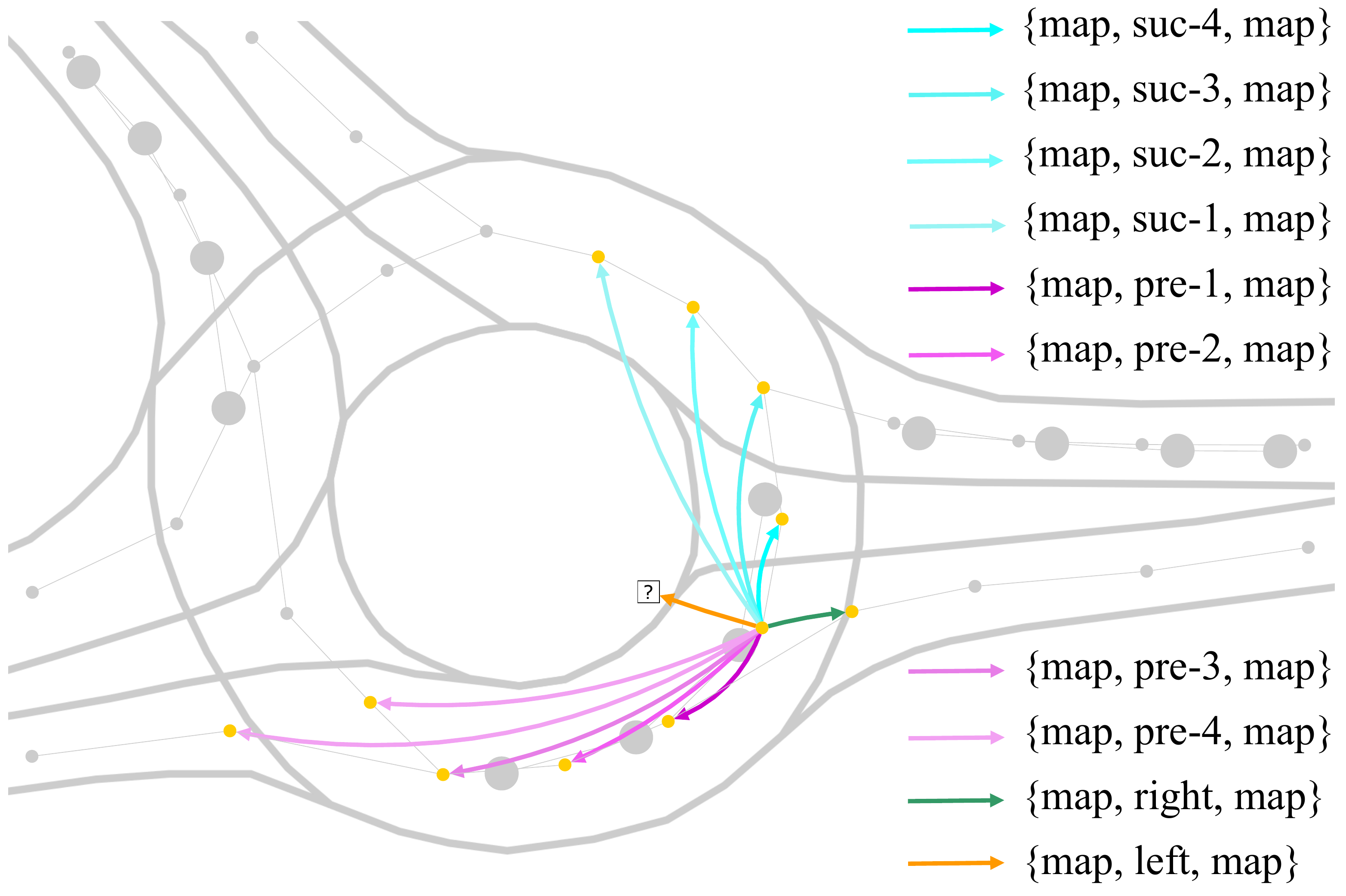}
        \caption{Edges among map-nodes. For a better view, only the edges from one map-node are depicted.}
    	\label{fig:map_edges}
	\end{subfigure}
	\begin{subfigure}{0.95\columnwidth}
    	\vspace{5pt}
    	\centering
    	\includegraphics[width=0.85\columnwidth]{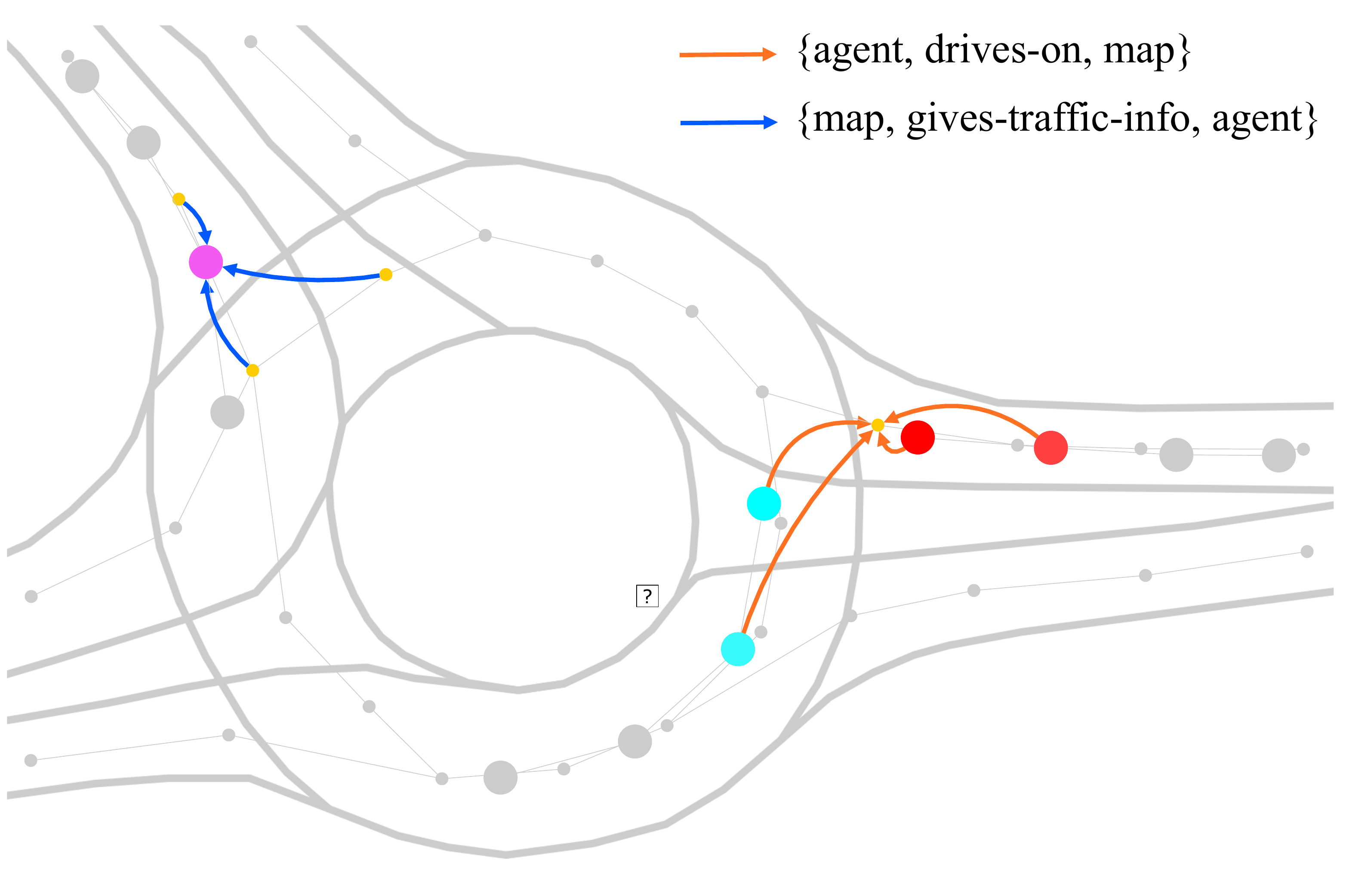}
        \caption{Edges between agent-nodes and map-nodes. For a better view, one agent-node is depicted as destination on the left and on the right one map-node is selected as destination.}
    	\label{fig:fusion_edges}
	\end{subfigure}
	\caption{Overview of the edges used by the heterogeneous GNN.}
	\label{fig:edges_overview}
\end{figure}
\subsection{Hetero GNN}
The heterogeneous graph is defined as $\mathcal{G = \{N,E}\}$, where $\mathcal{N}$ denotes the set of nodes and $\mathcal{E}$ denotes the set of edges with their corresponding edge features.
A scene consists of traffic participants, hereafter referred to as agents, and the HD-Map.
In this work the set of nodes $\mathcal{N = \{A,M\}}$ consist of two types:
\begin{itemize}
    \item The set of agent-nodes $\mathcal{A}$, where a single agent-node $\pmb{a}_i^t$ refers to a measurement at time-step $t$ of the observed past trajectory of the \textit{i}-th agent and consists of the current position, velocity and orientation, so that, $\pmb{a}_i = (x_i,y_i,v_{x_i},v_{y_i},h_i)^\intercal$.
    \item The set of map-nodes $\mathcal{M}$, where a single map-node $\pmb{m}_i$ refers to a segment of a centerline of the vectorized HD-Map consisting of direction and position, hence, $\pmb{m}_i = (x_i,y_i,\Delta x_i,\Delta y_i)^\intercal$.
\end{itemize}
The set of the different directed edge types $\mathcal{E}$ of the heterogeneous graph can be seen in Fig.~\ref{fig:edges_overview}.
The connections of a specific edge type from node type $j$ to node type $i$ with relation $r$ are stored in the adjacency matrix $\pmb{A}_{j,r,i}$ and $\pmb{e}_{j,r,i}$ denotes the corresponding edge features.
The edge features for each relation consist of the relative Cartesian coordinates between two connected nodes.

To update the node features $\pmb{x}_{i,r}^{(l)} \in \mathbb{R}^F$ of node $i$ in Layer $l$ for a specific edge type a basic message passing scheme introduced by Fey et al. \cite{torch_geometric} is used.
\begin{equation}
    \pmb{\hat{x}}_{i,r}^{(l)} = \gamma_{r}^{(l)} \left(\pmb{x}_i^{(l-1)}, \sum_{j\in \mathcal{N}(i)}\phi_{r}^{(l)}\left(\pmb{x}_i^{(l-1)},\pmb{x}_j^{(l-1)},\pmb{e}_{j,r,i}^{(l-1)}\right)\right)
    \label{eq:messagepassing}
\end{equation}
A MLP $\phi^{(l)}$ is used to calculate the messages of the neighboring nodes $\pmb{x}_j$ while also using the edge features $\pmb{e}_{j,r,i}$ from the edge connecting the corresponding node j to node i with relation $r$.
The neighboring nodes are determined by the associated adjacency matrix.
In Eq.~\ref{eq:messagepassing} the messages are aggregated using a sum.
The edge type specific update $\pmb{\hat{x}}_{i,r}^{(l)}$ is calculated by another MLP $\gamma^{(l)}$.
Afterwards all edge type specific node updates are merged by a sum followed by ReLU Activation, Residual-Connection and a Layer Normalization. to get the output of layer $l$:
\begin{equation}
    \pmb{x}_{i}^{(l)} = \text{norm}\left(\text{ReLU}\left(\sum_{r\in \mathcal{E}(i)} \pmb{\hat{x}}_{i,r}^{(l)}\right) + \pmb{x}_{i}^{(l-1)}\right)
    \label{eq:gnnlayer}
\end{equation}
In the proposed heterogeneous graph, not every edge is used simultaneously to update the nodes. Instead, edges between the same types of nodes are used first to generate more meaningful node features. Afterwards all edges in the graph except \{agent, merge, agent\} are used to further update node features and fuse context between agents and the map.\{agent, merge, agent\} is used to get the output of the GNN, i.e., a latent feature vector for each agent in the scene.
\subsubsection{Map Context}
\label{sec:mapcontext}
Map-nodes are connected to other map-nodes using the previous, successive, left and right neighbour according to the driving direction of the lane, e.g., \{map, left, map\} for the left neighbour.
During message passing it is preferable to propagate information along the road-direction rather than perpendicular to it because most road users travel along the road and not across.
We accomplish this by adding new edges along the road connecting a map node with its \textit{i}-th predecessor respectively successor using the \textit{i}-th power of the corresponding adjacency, e.g., \{map, pre-2, map\}.
A detailed view of the edges between map-nodes is given in Fig.~\ref{fig:map_edges} and is similar to LaneGCN \cite{lanegcn}.
For message generation we propose an extension to the basic GCN-Conv \cite{gcn}.
We include the usage of edge features in the message generation $\phi$ resulting in the node updates $\pmb{\hat{x}}_{i,r}^{(l)}$,
\begin{equation}
    \pmb{\hat{x}}_{i,r}^{(l)} = \sum_{j\in \mathcal{N}(i)} \frac{1}{\sqrt{\text{deg}(i)}  \sqrt{\text{deg}(j)}}  \left(  \left(\pmb{x}_{j}^{(l-1)} + \pmb{e}_{j,r,i}^{(l-1)}\right) \pmb{W} \right) + \pmb{b}
    \label{eq:edgeconv}
\end{equation}
where $\pmb{W}$ and $\pmb{b}$ refer to learnable parameters.
Edge and node features are added together, which reduces the number of learnable parameters without decreasing performance, see Seq.~\ref{sec:ablation}.
To gather a good encoding of the HD-Map Data we use five layers, where each layer is constructed according to Eq.~\ref{eq:gnnlayer}.
\subsubsection{Agent Context}
\label{sec:agentcontext}
An agent-node is connected to its predecessor and successor belonging to the past trajectory of the agent.
The corresponding edges are named \{\textit{agent, pre, agent}\} and \{\textit{agent, suc, agent}\}.
For social context \{\textit{agent, social, agent}\}, every agent-node is connected to agent-nodes of the previous, same and future timestamp that belong to other agents.
The respective edges are shown in Fig.~\ref{fig:agent_edges}.
Updating the agent-nodes is similar to the map-nodes.
In order to pass information from the first to the last agent-node of an agents trajectory, the number of used layers $n$ corresponds to the number of time-steps of the past trajectory, e.g., Argoverse: $n=20$, INTERACTION: $n=10$. Messages are generated using Eq.~\ref{eq:edgeconv}.
Social context is added during the last two layers with a multi head graph attention module (GATv2) \cite{gatv2}.
Therefore, edges of type \{\textit{agent, social, agent}\} are used.
The node updates via GATv2 \cite{gatv2} for relation $r$ are given by
\begin{equation}
    \pmb{\hat{x}}_{i,r}^{(l)} = \alpha_{i,i}^r \pmb{x}_i^{(l-1)}\pmb{W}_{1} + \sum_{j \in \mathcal{N}(i)} \alpha_{i,j}^r \pmb{x}_j^{(l-1)} \pmb{W}_{2}
\end{equation}
where the attention coefficients $\alpha_{i,j}$ are calculated with the learnable parameters $\pmb{w}$ and $\pmb{W}_{i \in\{1,2,3\}}$ as:
\begin{equation}
    \alpha_{i,j} = \text{softmax}\left(\text{LeakyReLU}\left( \left[\pmb{x}_{i} \mathbin\Vert \pmb{x}_{j} \mathbin\Vert \pmb{e}_{j,r,i}\right] \pmb{W}_{3} \right)\pmb{w}\right)
\end{equation}

\subsubsection{Context fusion}
To properly fuse the HD-Map with the past trajectories of the agents, we use edges of type \{\textit{agent, drives-on, map}\} and \{\textit{map, gives-traffic-info, agent}\}.
These two edges use a multi head GATv2 \cite{gatv2} module.
The source nodes are selected based on the euclidean distance $d_\text{th}$ of the 2d-position to the target nodes.
$d_\text{th}$ is dynamically calculated using the velocity of the agent-nodes and a threshold time $t_\text{th}$.
This compensates for faster moving agents.
Furthermore, we include the edges introduced in Seq.~\ref{sec:agentcontext} and Seq.~\ref{sec:mapcontext}. In total, we use two fusion layers.

The latent representation of an agents past trajectory is spread out between all agent-nodes belonging to this specific agent, hence, for every agent, the agent-node at $t_\text{obs}$ is selected as final feature vector and updated by a multi head GATv2 \cite{gatv2} module using edges from the past agent-nodes of that agent.
Fig.~\ref{fig:agent_edges} shows the edges \{\textit{agent, merge, agent}\} for this purpose.
\subsection{Motion-Prediction Head}
We use a combination of regression and scoring in separate MLPs to generate $K$ possible trajectories per agent.
For each mode, a new regression and classification MLP is instantiated.
Input is the latent feature vector for each agent.
To calculate the trajectory score we also use the predicted trajectory.
The two MLPs are similar and consist of a linear layer with a residual connection, ReLU Activation, Layer Normalization; followed by another linear layer.
The model outputs the predicted trajectories $\pmb{\mathcal{T}}$ of shape $[A,K,T_f,2]$ and the scores $\pmb{s}$ of shape $[A,K]$, where $A$ is the number of agents and $T_f$ equals the number of predicted time-steps.
\subsection{Loss}
The Loss $\mathcal{L}$ consists of a regression Loss and a classification Loss.
\begin{equation}
    \mathcal{L} = \mathcal{L}_{reg} + \lambda  \mathcal{L}_{cls}
\end{equation}
A smooth L1 Loss is used as regression Loss $\mathcal{L}_{reg}$.
To prevent mode collapse, $\mathcal{L}_{reg}$ is only calculated for the mode $k_{\text{min}}$ with minimal final displacement error (FDE) to the ground truth.
\begin{equation}
    \mathcal{L}_{reg} = \frac{1}{AT_f}\sum_{a}^{A}\sum_{t}^{T_f}\sum_{n \in\{x,y\}} \text{smoothL1}\left(\mathcal{T}_{a,n}^{t,k_{\text{min}}},\hat{\mathcal{T}}_{a,n}^{t}\right)
\end{equation}
with
\begin{equation}
    \text{smoothL1}(x,y) = 
    \begin{cases}
        0.5 * (x -y)^2, & \text{if} |x - y| < 1\\
        |x - y| - 0.5,              & \text{otherwise}
    \end{cases}
\end{equation}
where $\hat{\pmb{\mathcal{T}}}$ refers to the ground truth.
The classification loss $\mathcal{L}_{cls}$ is a max-margin loss \cite{multiclasshingeloss} with margin $m$.
\begin{equation}
    \mathcal{L}_{cls} = \frac{1}{A(K-1)} \sum_{a}^{A} \sum_{k \ne k_\text{min}}^{K} \max\left( 0,s_{a,k} + m - s_{a,k_\text{min}}\right)
\end{equation}
\section{Evaluation}
In the following, we evaluate our model on the INTERACTION dataset \cite{interaction} and the Argoverse motion forecast dataset \cite{argoverse}.
Firstly, we introduce the datasets, the evaluation metrics and the used hyperparameter settings.
Afterwards, we conduct ablation studies on the architecture and finally compare our model to the state-of-the-art.
\subsection{Experimental Settings}
The Argoverse Motion Forecast Dataset is a large scale collection of 323557 samples, each with a duration of $5~s$, resulting in a total of $320~h$.
The data was collected in Miami and Pittsburgh with $10~Hz$.
The task is to predict the future locations of one agent for $3~s$, given its history of $2~s$.
HD-Map data is provided in an argoverse specific format.

The INTERACTION Dataset is a highly interactive dataset recorded in 5 different locations including roundabouts, merging scenarios and intersections in Germany, USA and China.
It consists of around $16.5~h$ of data including 40054 trajectories sampled at $10~Hz$.
The task is to predict the future locations of all agents in the scene for $3~s$, given their history for $1~s$.
The HD-Map data is provided using the Lanelet2 \cite{lanelet2} format.

To evaluate the results quantitatively on Argoverse, we use its suggested metrics: Minimum Average Displacement Error (minADE), see Eq.~\ref{eq:minade}, Minimum Final Displacement Error (minFDE), see Eq.~\ref{eq:minfde}, and Minimum Miss Rate (minMR). Latter denotes the percentage of predictions having a minFDE greater than 2m. $\left\Vert \cdot \right\Vert^2$ refers to L2-Norm.
\begin{equation}
\text{minADE} = \frac{1}{AT_f} \sum_{a}^{A} \sum_{t}^{T_f} \min_{k}^{K} \left\Vert \mathcal{T}_{a}^{t,k} - \hat{\mathcal{T}}_{a}^{t} \right\Vert^2 %
\label{eq:minade}
\end{equation}
\begin{equation}
\text{minFDE} = \frac{1}{A} \sum_{a}^{A} \min_{k}^{K} \left\Vert \mathcal{T}_{a}^{T_f,k} - \hat{\mathcal{T}}_{a}^{T_f} \right\Vert^2 %
\label{eq:minfde}
\end{equation}
For multi-modal predictions, minFDE refers to the minimum euclidean distance of the predicted trajectory and the ground truth at the prediction horizon $T_{F}$ over all modes.
minADE is defined as the euclidean distance between the ground truth and the predicted positions averaged by time.
minMR indicates the ratio of the predictions where the final position of the trajectory of the best mode is more than a certain threshold, usually $2~m$, away from the ground truth.
On INTERACTION, we its proposed metrics Minimum Joint Average Displacement error (minJADE), see Eq.~\ref{eq:minjade}, Minimum Joint Final Displacement Error (minJFDE), see Eq.~\ref{eq:minjfde}, and Minimum Joint Miss Rate (minJMR) as metrics to measure the performance of joint motion prediction. Latter denotes the percentage of predictions having a minJFDE greater than 2m.
\begin{equation}
\text{minJADE} = \min_{k}^{K} \frac{1}{AT_f} \sum_{a}^{A} \sum_{t}^{T_f} \left\Vert \mathcal{T}_{a}^{t,k} - \hat{\mathcal{T}}_{a}^{t} \right\Vert^2 %
\label{eq:minjade}
\end{equation}
\begin{equation}
\text{minJFDE} = \min_{k}^{K} \frac{1}{A} \sum_{a}^{A} \left\Vert \mathcal{T}_{a}^{T_f,k} - \hat{\mathcal{T}}_{a}^{T_f} \right\Vert^2 %
\label{eq:minjfde}
\end{equation}
Training on each dataset was done on a RTX 3080 GPU for 40 epochs starting with an initial learning rate of 1e-3 and a decay of 0.5 every fifth epoch.
We used the Adam \cite{adam} optimizer, a batch size of 8 and a weight decay of 0.5\% for all weights which are not part of a normalization layer.
All attention modules have four heads and the results of the heads are concatenated.
Training took $33~h$ on Argoverse and $8~h$ on INTERACTION. For each scene in Argoverse \cite{argoverse}, we set the position of the last time-step of the ego-agent as the origin of the local fixed coordinate system.
In INTERACTION \cite{interaction} the origin of a sample is set to the geometric center point of all the trajectories in the sample.
For both datasets, we use a square with size of $160~m$ x $160~m$ centered around the origin to determine the relevant lanes and agents for the graph.
\subsection{Results}
Tab.~\ref{tab:sota} shows the results of our model in comparison to state-of-the-art approaches on the Argoverse and INTERACTION dataset.
We achieve similar results as the state-of-the-art while having only 2.5 Mio parameters.
In comparison to DenseTNT, HoliGraph predicts all traffic participants at once instead of only a single agent.
HoliGraph has an inference time of around $77~ms$ on a RTX 3080, making it real-time capable.
It can be assumed that the performance could be further increased by adding more semantic features to the graph, e.g., differentiate between road-bound and non-road-bound agents and adding traffic lights to the map-nodes.
\begin{table}[h]
    \centering
    \hspace{0.1pt}
    \caption{Results on argoverse test, regular INTERACTION single and regular INTERACTION multi test dataset.}
    \begin{tabular}{l | c c c | c}
        \hline
        argoverse & \multicolumn{3}{c|}{K=6} & No. of  \\ 
        single & minADE & minFDE & minMR & Parameters\\ \hline
        HoliGraph (ours) & 0.98~m & 1.65~m & 0.172 & 2.5 Mio\\ 
        DenseTNT \cite{densetnt}& 0.94~m & 1.49~m & 0.105 & 1.4 Mio\\
        Scene Trans.\cite{scene_transformer} & 0.80~m & 1.23~m & 0.13 & 15.3 Mio\\
        LaneGCN \cite{lanegcn} & 0.87~m & 1.36~m & 0.16 & 3.6 Mio\\ \hline \hline
        interaction & \multicolumn{3}{c|}{K=6} & No. of \\ 
        single & minADE & minFDE & minMR & Parameters\\ \hline
        HoliGraph (ours) & 0.213~m & 0.529~m & 0.029 & 2.5 Mio\\ 
        DenseTNT \cite{densetnt} & 0.2819~m & 0.6371~m & 0.028 & 1.4 Mio\\
        HDGT \cite{hdgt} & 0.1085~m & 0.3361~m & 0.014 & 12 Mio\\ \hline \hline
        interaction & \multicolumn{3}{c|}{K=6}  & No. of\\ 
        multi & minJADE & minJFDE & minJMR & Parameters \\ \hline
        HoliGraph (ours) & 0.362~m & 1.043~m & 0.138 & 2.5 Mio \\ 
        ReCoG2 \cite{recog} & 0.330~m & 0.932~m & 0.194 & -\\ 
        HDGT \cite{hdgt} & 0.2162~m & 0.7309~m & 0.1384 & 12 Mio\\ \hline
    \end{tabular}
    \label{tab:sota}
\end{table}
\begin{figure*}[t]
    \centering
    \vspace{10pt}
    \begin{subfigure}{0.49\textwidth}
        \centering
        \includegraphics[width=\columnwidth]{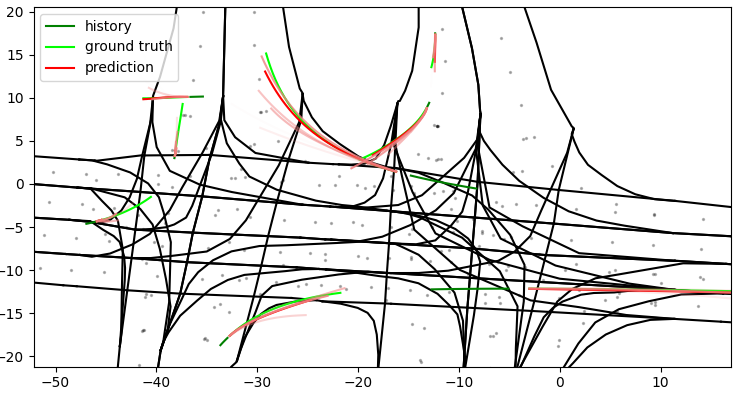}
        \caption{Dense traffic scene}
        \label{fig:dense_scene}
    \end{subfigure}
    \begin{subfigure}{0.49\textwidth}
        \centering
        \includegraphics[width=\columnwidth]{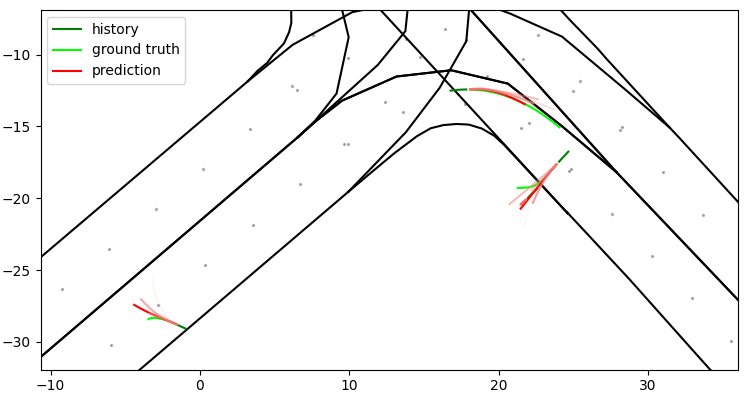}
        \caption{Pedestrian crossing}
        \label{fig:pedestrian}
    \end{subfigure}
    \caption{Qualitative results on INTERACTION validation dataset. History is depicted in dark green, ground truth in light green, predictions in light red. The prediction with the highest score is depicted in red. The map-nodes are depicted as light red dots.}
    \label{fig:results}
\end{figure*}
\addtolength{\textheight}{+0cm}   %
\subsection{Ablation Study}
\label{sec:ablation}
To investigate the effect of using different types of context information on the prediction accuracy, we conducted an ablation study.
Tab.~\ref{tab:context} shows that the performance on the INTERACTION validation dataset is increasing when providing the model with more context information.
It also shows the importance of edge features to provide the model with additional relational information.
\begin{table}[h]
    \hspace{0.1pt}
    \caption{Results on INTERACTION validation dataset for different types of context information as input. History means each agent is only connected to itself. Map means the usage of map-nodes. Social means that agents are also connected to other agents. Relational refers to the usage of edge features.}
    \centering
    \begin{tabular}{>{\centering}m{0.6cm}
                    >{\centering}m{0.4cm}
                    >{\centering}m{0.6cm}
                    >{\centering}m{1.0cm} |
                    >{\centering}m{1.0cm}
                    >{\centering}m{1.0cm}
                    >{\centering\arraybackslash}m{0.9cm}}
        \hline
        \multicolumn{4}{c|}{context information} & \multicolumn{3}{c}{K=6}  \\ 
        history & map & social & relational & minJADE & minJFDE & minJMR \\ \hline
        \checkmark & & & & 0.607~m & 1.745~m & 0.311 \\
        \checkmark & & \checkmark & & 0.562~m & 1.601~m & 0.272 \\
        \checkmark & \checkmark & & & 0.458~m & 1.254~m & 0.188 \\
        \checkmark & \checkmark & \checkmark & & 0.441~m & 1.212~m  & 0.178 \\ 
        \checkmark & \checkmark & \checkmark & \checkmark & 0.362~m & 1.043~m & 0.138 \\ \hline
    \end{tabular}
    \label{tab:context}
\end{table}
In Tab.~\ref{tab:architecture} we investigate the effect of residual connections during node update, the temporal encoding of agent-nodes and the way of including edge features.
The residual connections improve the model performance by $11~\%$ (mean over all metrics).
Adding timestamp information directly to the agent-nodes with the temporal encoding from Eq.~\ref{eq:temporal} further improves performance by $6~\%$ (mean over all metrics).
The third row in Tab.~\ref{tab:architecture} refers to the architecture used in the final model which does not use the concatenation of edge features and node features. Thats because the concatenation only results in a small performance gain, but significantly increases the number of learnable parameters from 2.5 Mio to 4.2 Mio.
\begin{table}[h]
    \caption{Results on INTERACTION validation dataset for different architectures. Residual means residual connections during node update. Temporal refers to the temporal encoding of agent-nodes and concat refers to the concatenation of edge and node features.}
    \centering
    \begin{tabular}{c c c | c c c}
        \hline
        \multicolumn{3}{c|}{architecture} & \multicolumn{3}{c}{K=6}  \\ 
        residual  & temp & concat & minJADE & minJFDE & minJMR \\ \hline
         & & & 0.426~m & 1.214~m & 0.177 \\
        \checkmark & & & 0.383~m & 1.090~m & 0.151 \\
        \checkmark & \checkmark & & 0.362~m & 1.043~m & 0.138 \\
        \checkmark & \checkmark & \checkmark & 0.361~m & 1.039~m & 0.137 \\ \hline
    \end{tabular}
    \label{tab:architecture}
\end{table}
Lastly we investigate the influence of different attention mechanisms. All three modules result in roughly the same number of learnable parameters, but the GATv2 module outperforms the other attention-modules.
\subsection{Qualitative Results}
Some qualitative results are depicted in Fig.~\ref{fig:results}.
Fig.~\ref{fig:dense_scene} shows the performance of the model in a complex intersection with a lot of interactions between the traffic participants.
In the scene are vehicles as well as pedestrians present.
Our model is able to predict all agents well.
For most agents, the lateral predictions are almost perfect. However, their longitudinal predictions show small deviations.
On the right side of Fig.~\ref{fig:pedestrian} a pedestrian is crossing the road. Nearly all modes indicate a light left turn.
This is a result of the attention to the map-nodes, as the driving direction of the road is to the right. We will use this showcase as a motivation to distinguish in our future work between road-bound and non-road-bound users.
\begin{table}[h]
    \centering
    \caption{Results on INTERACTION validation dataset for different attention mechanism.}
    \begin{tabular}{c c c | c c c}
        \hline
        \multicolumn{3}{c|}{attention modules} & \multicolumn{3}{c}{K=6}  \\ 
        GAT & GATv2 & Transformer & minJADE & minJFDE & minJMR \\ 
        \cite{gat} & \cite{gatv2} &  \cite{transformer_gat} & & &\\ \hline
        \checkmark & & & 0.434~m & 1.207~m & 0.178 \\
         & \checkmark & & 0.362~m & 1.043~m & 0.138 \\
         & & \checkmark & 0.416~m & 1.149~m & 0.163 \\ \hline
    \end{tabular}
    \label{tab:attention}
\end{table}
\section{Conclusion}
In this paper we have proposed a new way to represent temporal information in heterogeneous graphs for motion prediction.
Instead of compressing the temporal information, we embed the whole past trajectories of all agents into the GNN.
We achieve similar results as state-of-the-art approaches while having considerably less learnable parameters.
We did an extensive ablation study to verify the effectiveness of each design decision.
The evaluation was conducted on two different state-of-the-art datasets.
As the holistic graph representation allows to include arbitrary information, we are going to further distinguish between road-bound agents such as cars, trucks and motorcycles and non-road-bound agents such as pedestrians.

Additionally, we plan to investigate the suitability of the HoliGraph representation for tracking tasks. The main problem of associating detected objects to tracks could be solved by learning the most probable edges, i.e., \{agent, pre, agent\}, connecting corresponding agent-nodes of the same track.
\section*{ACKNOWLEDGMENT}
The research leading to these results was conducted within the project KIsSME (Artificial Intelligence for selective near-real-time recordings of scenario and maneuver data in testing highly automated vehicles) and was funded by the German Federal Ministry for Economic Affairs and Climate Action.
Responsibility for the information and views set out in this publication lies entirely with the authors.
\printbibliography
\end{document}